\documentclass[runningheads]{llncs}
\usepackage{graphicx}
\usepackage{tikz}
\usepackage{comment}
\usepackage{amsmath,amssymb} % define this before the line numbering.
\usepackage{color}
\usepackage{multirow}
\usepackage{subfigure}

\graphicspath{{./figures/}}

\begin{document}
% \renewcommand\thelinenumber{\color[rgb]{0.2,0.5,0.8}\normalfont\sffamily\scriptsize\arabic{linenumber}\color[rgb]{0,0,0}}
% \renewcommand\makeLineNumber {\hss\thelinenumber\ \hspace{6mm} \rlap{\hskip\textwidth\ \hspace{6.5mm}\thelinenumber}}
% \linenumbers
\pagestyle{headings}
\mainmatter
\def\ECCVSubNumber{}  % Insert your submission number here

\title{Bi-Dimensional Feature Alignment for Cross-Domain Object Detection} % Replace with your title

% INITIAL SUBMISSION 
\begin{comment}
\titlerunning{ECCV-20 submission ID \ECCVSubNumber} 
\authorrunning{ECCV-20 submission ID \ECCVSubNumber} 
\author{Anonymous ECCV workshop submission}
\institute{Paper ID \ECCVSubNumber}
\end{comment}
%******************

% CAMERA READY SUBMISSION
%\begin{comment}
\titlerunning{Bi-Dimensional Feature Alignment}
% If the paper title is too long for the running head, you can set
% an abbreviated paper title here
%
\author{Zhen Zhao\inst{1} \and
Yuhong Guo\inst{1,2} \and
Jieping Ye\inst{1} 
}
\authorrunning{Z. Zhao et al.}
% First names are abbreviated in the running head.
% If there are more than two authors, 'et al.' is used.
%
\institute{$^1$\; DiDi Chuxing \qquad\quad
$^2$\; Carleton University 
%{\{alexzhaozhen,yejieping\}@didiglobal.com},\;
%{yuhong.guo@carleton.ca}
}
%\end{comment}
%******************
\maketitle

\begin{abstract}
Recently the problem of cross-domain object detection has started drawing attention in the computer vision community.
In this paper, we propose a novel unsupervised cross-domain detection model that exploits the annotated 
data in a source domain to train an object detector for a different target domain. 
The proposed model mitigates the cross-domain representation divergence for object detection
by performing cross-domain feature alignment in two dimensions,
the depth dimension and the spatial dimension. 
In the depth dimension of channel layers, it uses 
inter-channel information 
to bridge the domain divergence with respect to image style alignment. 
In the dimension of spatial layers, it deploys spatial attention modules
to enhance detection relevant regions and suppress irrelevant regions
with respect to cross-domain feature alignment. 
Experiments are conducted on a number of benchmark cross-domain detection datasets.
The empirical results show the proposed method outperforms the state-of-the-art comparison methods. 

\end{abstract}

\section{Introduction}

\begin{figure}[t]
\centering
\vskip .1in
\includegraphics[width=1.0\linewidth]{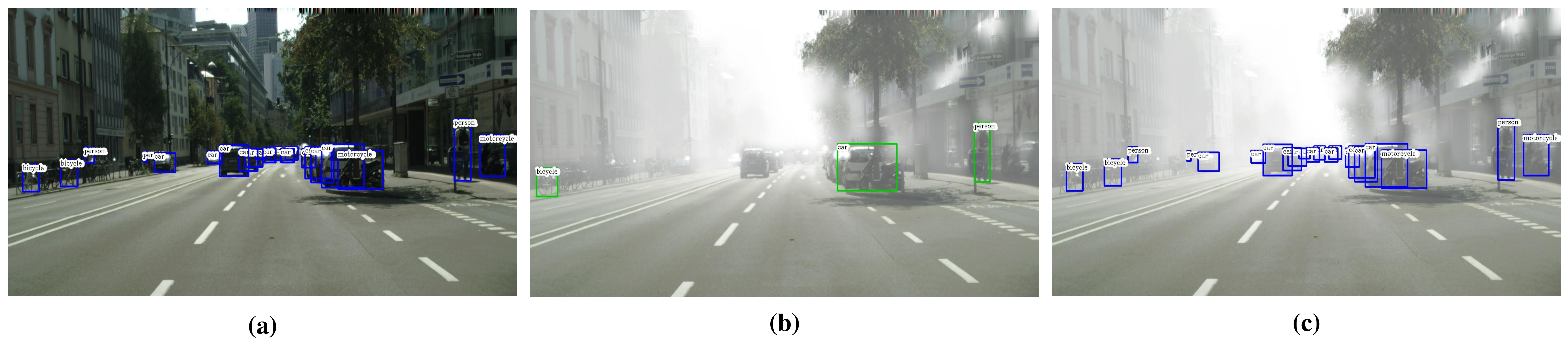}
\caption{Example of deployment of supervised object detector with different training and test datasets, Cityscapes and Foggy Cityscapes. 
	(a) Labeled image example from the training set, Cityscapes. 
	(b) The detection result on an image in Foggy Cityscapes 
	using the detector trained on Cityscapes. 
	(c) The ground-truth annotation of the Foggy Cityscapes example. 
	This example shows that weather-induced domain gaps can lead to performance degradation.}
\label{fig:one}
\end{figure}

The deployment of supervised deep learning models has led to great advance
in many computer vision tasks such as image classification~\cite{simonyan2014very}, 
object detection~\cite{girshick2014rich,girshick2015fast,ren2015faster,liu2016ssd}, 
and image segmentation~\cite{Zhao_2017_CVPR}.
However, their success relies on the assumptions of standard supervised learning;
that is, the deep models need to be trained with a sufficient amount of i.i.d. labeled samples 
that come from the same distribution as the test data.
In practice, due to factors such as the collection means or weather conditions, 
the operational test dataset can be different from the training dataset,
which can significantly degrade the performance of image analysis systems.
For example, Fig.~\ref{fig:one} presents the direct deployment result of an object detector trained in 
one domain, Cityscapes, and applied in another domain, Foggy Cityscapes.
It shows the detection model trained with images collected in normal weather 
fails to detect many objects on images collected in foggy weather.
Although one can solve this problem by collecting labeled data from the same test dataset,
the data annotation/labeling process is typically time-consuming and expensive.
To avoid the expensive needs of repeatedly collecting labeled images,
many unsupervised domain adaptation methods have been developed 
for image segmentation and classification tasks to overcome the cross-domain performance degradation
\cite{ganin2016domain,long2018conditional,cicek2019unsupervised,zhang2017curriculum,tsai2018learning,tsai2019domain}. 
However, much less effort has been devoted to the more complex cross-domain object detection task. 

As a detector needs to identify both the objects and their precise locations in an image, 
it is more challenging to design an effective cross-domain detector than a cross-domain classifier.
One early work for adaptive object detection 
\cite{chen2018domain} adopts a domain adversarial feature alignment strategy at both
the global image level and the proposal instance level. 
However, 
global image feature alignment is more effective to domain shifts over image appearances and textures,
while unsuitable for handling cross-domain spatial distribution divergences. 
A more recent work~\cite{saito2019strong} improves adaptive detection
by deploying strong cross-domain alignment at low-level features such as local textures/colors
and weak alignment at high-level global image features.
Nevertheless, this work still fails to explore the spatial properties of features 
which are essential for object detection.

In this paper, we propose a novel end-to-end deep learning model for 
cross-domain object detection by aligning features from the source and target domains
in both the depth and spatial dimensions. 
Our assumption is that the image representation can be captured from the perspectives of
both the semantic contents (e.g., the objects contained in the image) and the style of the image,
and hence cross-domain feature alignment should be addressed from both aspects.
Following previous work~\cite{Gatys2016Image}, we represent the style of an image
using the inter-channel Gram matrix computed over features in the depth dimension of the feature map,
which captures the correlations between the different filter responses along the spatial dimension,
and can be adversarially aligned across domains.
For cross-domain content feature alignment, we propose to use an attention module 
along the spatial dimension
to enhance features in important regions (e.g., regions with objects)
and suppress features in irrelevant background areas. 
This attention module not only will guide the domain adaption model to form 
a region-sensitive domain adversarial feature alignment, 
but also will be added into the feature representations 
of the backbone network to facilitate the consequent region proposal and local object classification steps of the detector.
Overall the contribution of this work can be summarized as follows:
(1) This is the first domain adaptation work that performs cross domain 
semantic content and style feature alignments separately and simultaneously 
in the spatial and depth dimensions. 
(2) We deploy a novel spatial attention module to achieve target region sensitive cross-domain feature alignment.
(3) We conduct extensive experiments on benchmark cross-domain detection datasets and
the proposed model achieves the state-of-the-art performance.

%%%%%%%%%%%%%%%%%%%%%%%%%%%%
\section{Related Work}
\noindent{\bf Object Detection.} 
The development of convolutional neural networks (CNN) has led to great advance in object detection.
Traditional object detection methods use sliding windows and manual feature classification designs~\cite{Navneet2005H,Felzenszwalb2010Object}. 
In recent years, a two-stage detection strategy based on region of interest (ROI) 
has gained wide applicability~\cite{girshick2014rich,girshick2015fast,ren2015faster}. 
The early RCNN model~\cite{girshick2014rich} uses selective search to generates a set of region proposals for object detection. 
Fast-RCNN~\cite{girshick2015fast} improves RCNN by identifying region proposals and deploying ROI pooling on the convolutional feature map of CNN.
Faster-RCNN \cite{ren2015faster} 
combines Region Proposal Network (RPN) and Fast-RCNN to replace the previous selective search and 
further improve the detection performance. 
As a landmark detection model, Faster-RCNN provides the basis for many subsequent research studies~\cite{liu2016ssd,Redmon2016You,lin2017feature,he2017mask}. 
This paper and many related unsupervised domain adaptive object detection methods 
also use Faster-RCNN as the backbone detection model.\\

\noindent{\bf Unsupervised Domain Adaptation.} 
Unsupervised domain adaptation, which aims to train a model in a label-rich source domain 
for using in an unlabeled target domain, has attracted a lot of attention in the computer vision community. 
Many works have tried to learn cross-domain aligned feature representations
\cite{dziugaite2015training,ganin2016domain,shen2017wasserstein,long2018conditional,choi2019self}. 
The work in~\cite{ganin2016domain} used a gradient reversal layer (GRL) to achieve the adversarial feature alignment operation.
The authors of \cite{long2018conditional} proposed a
conditional adversarial domain adaptive method by using category predictions as an additional input for the domain discriminator. 
The work in~\cite{choi2019self} used an image generation mechanism to achieve cross-domain transformation through image pixel-level alignment. 
There are also some studies that perform domain adaptation by minimizing various feature distribution distances between different domains,
such as 
the maximum mean discrepancy(MMD)~\cite{dziugaite2015training} and the Wasserstein distance~\cite{shen2017wasserstein}. 
However, most of these studies focus on image classification and segmentation tasks.\\

\noindent{\bf Domain Adaptation for Object Detection.} 
Although there are many works on cross-domain image classification and segmentation, 
domain adaption for object detection has just begun to receive attention. 
One relatively early work~\cite{chen2018domain} proposed to align image-level features and 
instance-level features with adversarial domain adaptation strategy for adaptive object detection.
The work in~\cite{inoue2018cross} used image pixel-level transitions and pseudo-labeling 
to achieve cross-domain weakly supervised target detection.
Another work~\cite{kim2019diversify} used Cycle-GAN~\cite{zhu2017unpaired} to generate multiple intermediate domain images between 
the source domain and the target domain to learn domain invariant representations.
The work~\cite{saito2019strong} proposed a multi-level adversarial feature alignment strategy, 
global weak alignment and local strong alignment, to improve cross-domain detection performance.
Multi-level alignment is also adopted in~\cite{xie2019multi},
which aligns the distributions of local features and global features simultaneously. 
while another work~\cite{zhu2019adapting} focused on selective alignment of related areas.
The work~\cite{Zhaoetal20} performs conditional adversarial global feature alignment
with dual multi-label prediction.
The authors of~\cite{HeMulti} adopted the idea of layered alignment by adding proportional reduction and 
weighted gradient inversion layers to achieve domain invariance.
Different from these existing methods, our proposed approach induce domain invariant features
by enforcing not only multi-level alignments, but also multi-dimensional alignments.

%%%%%%%%%%%%%%%%%%%%%%%%
\begin{figure}[t]
\vskip .1in	
\begin{center}
\includegraphics[width=0.96\linewidth]{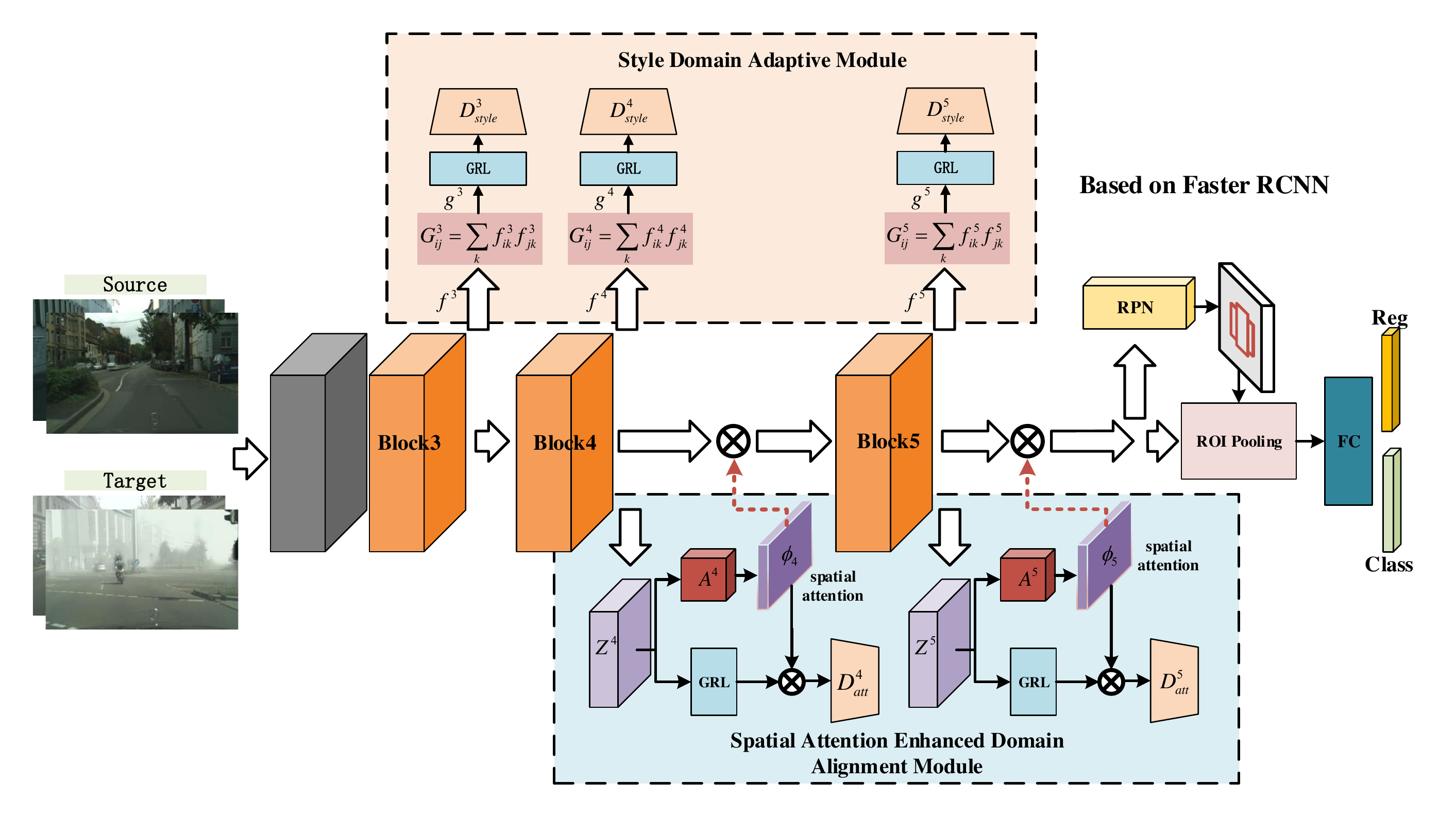}
\end{center}
	\caption{The model structure of the proposed multi-dimensional domain adaptive object detection method, 
	the Style and Spatial Attention enhanced feature alignment method for Domain Adaptive detection (SSA-DA). 
	The proposed model uses Faster-RCNN as its backbone network and has two major adaptive components: 
	the style domain adaptive module and the spatial attention enhanced domain alignment module. 
	}
\label{fig:two}
\end{figure}
%%%%%%%%%%%%%%%%%%%%%%%%
\section{Methodology}
In this section, 
we consider the unsupervised cross-domain object detection problem, 
where the source domain contains fully labeled data
and the data in the target domain is entirely unannotated.
Let $X_s$ denote the fully annotated source domain data, such that $X_s=\{(x_i^s, {\bf b}_i^s, {\bf c}_i^s)\}_{i = 1}^{{n_s}}$, 
where $x_i^s$ denotes the $i$-th image, ${\bf b}_i^s$ and ${\bf c}_i^s$ represent the bounding box coordinates and the corresponding labels respectively for 
the objects contained in the $i$-th image. 
Let $X_t=\{x_i^t\}_{i = 1}^{{n_s}}$ represent the unannotated images from the target domain. 
We aim to develop a good cross-domain detection method
that trains an object detector on these available image resources 
to perform well in the target domain.

In this work we propose a bi-dimensional feature alignment method, 
a Style and Spatial Attention enhanced feature alignment method for Domain Adaptive detection (SSA-DA).
The main idea of SSA-DA is to model the cross-domain style representation divergence and
semantic content representation divergence separately by
aligning image styles across domains in the depth dimension 
with style domain adaptive modules
and aligning detection effective features in the spatial dimension with spatial attention enhanced domain alignment modules. 
SSA-DA adopts the widely used Faster-RCNN as the backbone detection network, while
the feature alignments can be conducted in multiple layers of the feature extraction subnetwork. 
The overall structure of the proposed SSA-DA model is illustrated in Fig.~\ref{fig:two}.
As shown in Fig.~\ref{fig:two}, a style domain adaptive module is 
added after the 3rd, 4th, and 5th convolution blocks separately,
and a spatial attention enhanced domain alignment module is added
after the 4th and 5th convolution blocks separately.
The details of the two types of adaptive alignment modules 
and the overall learning problem will be introduced below.

\subsection{Depthwise Style Domain Adaptive Module}\label{section:style}

Image style is an important aspect of the image representation.
Variations in image styles can hinder the cross-domain object detection performance. 
Following a previous work~\cite{Gatys2016Image}, 
we build the style representation of an image in a feature space that captures texture information,
by calculating inter-channel feature correlations, i.e., correlations between
different filter responses,  in the feature map produced in any layer of the feature extraction network.
Let 
the feature map obtained for a given image $x$ after the $l$-th convolution block in the backbone of Faster-RCNN 
be expressed as $Z^l=F^l(x)\in{\mathbb{R}^{C^l\times H^l\times W^l}}$,
where $C^l$ denotes the number of channels (i.e., filters), 
$H^l$ and $W^l$ denote the spatial dimensions. 
Each channel contains the responses of the corresponding convolution filter of the current layer. 
To facilitate calculation, 
we transform $Z^l$ into a two-dimensional matrix $f^l\in {\mathbb{R}^{C^l \times M^l}}$ with $M^l=H^l\cdot W^l$,
such that the rows and columns of the matrix represent the channel and spatial dimensions respectively.
Then the style features can be calculated 
as a Gram matrix ${G}^l \in {\mathbb{R}^{C \times C}}$, such that
\begin{gather}
{G^l_{ij}} = \sum\limits_k {f_{ik}^l f_{jk}^l}
\end{gather}
Each entry $G^l_{ij}$ captures the inter-channel correlations between the $i$-th and $j$-th channels.
For simplicity, we can further reshape $G^l$ into a vector form $g^l\in\mathbb{R}^{C^{l2}\times 1}$,
which contains the style features produced. Such style features capture the texture information
but not the location arrangement.

To overcome the cross-domain style variation, we propose to align the style features across domains at a given layer
with an adversarial domain adaptation mechanism based on 
the generative adversarial network (GAN), 
which can effectively align two distributions
\cite{goodfellow2014generative}. 
As shown in Fig.~\ref{fig:two}, a style domain adaptive module is used to induce style-invariant feature representations
at multiple convolution blocks. 
At the $l$-th block, a domain discriminator 
$D^l_{style}$ is introduced to predict the domain of an input image style feature vector $g^l$, 
with $D^l_{style}(g^l)$ denoting the predicted probability of $g^l$ coming from the source domain. 
The feature alignment can be achieved through a min-max game between
the feature extractor $F^l$ and the domain discriminator $D^l_{style}$: 
\begin{align}
\label{eq2}
	\min_{F^l} \max_{D^l_{style}} &\quad   L_{style}^l =   \frac{1}{2}(L{_{style}^{ls}} + L{_{style}^{lt}})
\\
	L{_{style}^{ls}} = &  \quad \mathbb{E}_{x_s\in X_s}\left[ {{{(1 - D_{style}^l(g_s^l))}^\gamma }\log (D_{style}^l(g_s^l))}\right]
\\
	L{_{style}^{lt}} = &  \quad \mathbb{E}_{x_t\in X_t} \left[{{{(D_{style}^l(g_t^l))}^\gamma }\log (1-D_{style}^l(g_t^l))}\right]
\end{align}
Here we adopted the focal loss~\cite{lin2017focal,saito2019strong} in this adversarial training objective
to give more weights to examples that are hard to classify by $D^l_{style}$,
and $\gamma$ is a modulation factor that controls the contribution degree of the hard examples.
$g_s^l$ represents the style feature vector generated 
from the convolutional feature map $F_l(x_s)$ from the source domain 
and $g_t^l$ represents the style feature vector generated from $F_l(x_t)$
from the target domain. 
In the adversarial min-max game, the discriminator $D^l_{style}$ tries to maximally
discriminate the source domain features from the target domain features,
where the feature extractor $F^l$ tries to induces style features such that
the discriminator can be maximally confused.  

The style representation is a multi-scale representation involving multiple layers of the deep network.
The style features obtained from lower level layers reflect more pixel level information,
while the style features from higher level layers reflect more image structural information
\cite{Gatys2016Image,JohnsonPerceptual}. 
Therefore, we apply adversarial style feature alignment on multiple convolution blocks 
to obtain stable and multi-scale style domain adaptation of image features. 
In particular, in the proposed model, one style domain adaptive module is 
added to each of the convolution block 3, block 4 and block 5 of the backbone network of 
the Faster-RCNN before the region proposal network (RPN).
The overall multi-level style adversarial training loss $L_{style}$ can be summarized as follows:
\begin{align}
\label{eq5}
{L_{style}} = \sum\nolimits_{l = 3}^5 {\mathop {\min }\limits_{{F^l}} \mathop {\max }\limits_{D_{style}^l} L_{style}^l}
\end{align}
%

%%%%%%%%%%%%%%%%%%%%%%%%%%%%%%%%%%%%%%%%%%%%%%%%%%%%%
\subsection{Spatial Attention Domain Alignment Module}

Image features that reflect the semantic contents of images
are essential for object detection and it is important to bridge domain divergence 
over the image content features.
In general, an object is usually localized into some local region in an image
and the detector only needs to recognize the relevant regions.
Therefore directly aligning the global image features across domains might not 
be the most suitable strategy as these features can be dominated by irrelevant regions,
especially when the objects are small.
To address this problem, we propose a spatial attention domain alignment module,
as shown in Fig.~\ref{fig:two},  
which learns spatial attention to enhance features from the semantically relevant regions. 
Specifically, given the feature map produced from the $l$-th convolution block, $Z^l=F^l(x)$, 
we follow the method of~\cite{WooCBAM} to generate a spatial attention map $\phi^l\in\mathbb{R}^{1\times H^l\times W^l}$ 
using an attention network $A^l$, such that 
$\phi^l = {A}^l(Z^l)$, where ${A}^l$ is 
a convolution network with filters of $7 \times7$.
Then the spatial attention enhanced feature map can be produced as 
$Z^l_\phi$=$\phi^l \otimes Z^l$, where $\otimes$ denotes a replicated element-wise product operator 
that multiplies $\phi^l$ to each channel of $Z^l$.
It is expected that the learned spatial attention can enhance features in relevant regions of $Z^l$
and diminish features in irrelevant regions.

Given the spatial attention enhanced feature map $Z^l_\phi$, we introduce a domain discriminator 
$D_{att}^l$ to perform adversarial cross-domain feature alignment. 
Similar to the adversarial alignment on style features, we use focal loss in the adversarial objective, 
such that the min-max adversarial optimization problem is formulated as:
\begin{align}
\label{eq6}
	\mathop {\min }\limits_{{F^l}} \mathop {\max }\limits_{D_{att}^l} & 
	\quad L_{att}^l =  \frac{1}{2}(L{_{att}^{ls}} + L{_{att}^{lt}})
\\	
	L{_{att}^{ls}} \;= & \quad \mathbb{E}_{x_s\in X_s} \left[ {{{(1 - D_{att}^l(Z{{_\phi ^l}_s}))}^\varepsilon }\log (D_{att}^l(Z{{_\phi ^l}_s}))}\right]
\\
	L{_{att}^{lt}} \;= & \quad \mathbb{E}_{x_t\in X_t} \left[{{{(D_{att}^l(Z{{_\phi ^l}_t}))}^\varepsilon }\log (1 - D_{att}^l(Z{{_\phi ^l}_t}))}\right]
\end{align}
where $\varepsilon$ is a modulation factor hyperparameter for the focal loss; %
${Z_\phi ^l}_s$ and ${Z_\phi ^l}_t$ are the spatial attention enhanced feature maps from the $l$-th block 
for the source domain image $x_s$ and the target domain image $x_t$ respectively. 
Moreover, 
as previous work~\cite{Gatys2016Image,JohnsonPerceptual} has shown 
effective semantic information that can characterize image content is often available 
in the deep layers of a convolutional network instead of the shallow layers,
we propose to perform multi-level spatial attention enhanced domain alignment
on the last two convolution blocks of the Faster-RCNN, i.e., block 4 and block 5 in Fig.~\ref{fig:two}.  
The overall spatial attention enhanced adversarial training loss $L_{att}$ can be written as follows:
\begin{align}
\label{eq10}
{L_{att}} = \sum\nolimits_{l = 4}^5 {\mathop {\min }\limits_{{F^l}} \mathop {\max }\limits_{D_{att}^l} L_{att}^l}
\end{align}
In this min-max adversarial training, 
the feature extraction network $\{F^4, F^5\}$ will tries to maximally confuse the 
domain discriminators 
$\{D_{att}^4, D_{att}^5\}$ and align the spatial attention enhanced features across domains.

Meanwhile, the spatial attention enhanced feature map $Z^l_\phi$ at each block will be used as the input for the next block
along the backbone of the detection network. 
The attention enhanced feature map, ${Z^5_\phi}$, 
at the last convolution block, block 5, will be provided to the region proposal network (RPN) 
to produce region proposals and perform object classification and bounding box regression. 
The object detection loss $L_{det}$ in the source domain can be expressed as:
\begin{gather}
\begin{split}
	{L_{det}} =  \frac{1}{{{n_s}}}\sum\nolimits_{i = 1}^{{n_s}} {{L_{cr}}}(R(Z^{5}_{\phi_{i}} ),({\bf b}_i^s,{\bf c}_i^s))
\end{split}
\end{gather}
where $R$ denotes the combined function for the RPN, region classification and regression modules
of the Faster-RCNN, and $L_{cr}$ represents all the supervised classification loss and regression loss.

%%%%%%%%%%%%%%%%%%%%%%%%%%%%%%%%%%%%%%%%

\subsection{Overall Adversarial Learning}
We combine object detection loss $L_{det}$, style adversarial training loss $L_{style}$, 
and the spatial attention enhanced adversarial training loss $L_{att}$ together to 
form the following overall adversarial learning objective:
\begin{align}
{L_{all}} = {L_{det}} + \lambda {L_{style}} + \mu {L_{att}}
\label{eq11}
\end{align}
where $\lambda$ and $\mu$ are the trade-off parameters to balance different loss terms. 
This overall learning problem minimizes the detection loss on the labeled source domain data,
while bridging the representation gap between the source and target domains from both the style 
and content perspectives.
SGD optimization algorithm is used to perform training, while
GRL~\cite{ganin2016domain} is adopted to implement the gradient sign flip for the domain discriminator update.

%%%%%%%%%%%%%%%%%%%%%%%%%%%%%%%%%%%%%%%%

\section{Experiments}

To evaluate the proposed SSA-DA model, we conducted experiments 
on benchmark cross-domain detection tasks in three cross-domain variation scenarios: 
(1) {Normal to foggy weather variation}. 
In this scenario, we used the cross-domain detection task of adapting from
Cityscapes~\cite{cordts2016cityscapes} to Foggy Cityscapes~\cite{sakaridis2018semantic}.
(2) {Virtual to real scene variation}. The adaptive detection task from SIM-10K~\cite{JohnsonDriving} to Cityscapes~\cite{cordts2016cityscapes}
is used in this scenario. 
(3) {Cross-camera situation}. 
We used data collected with two different cameras to form the cross-domain detection task from KITTI~\cite{geiger2012we} and Cityscapes~\cite{cordts2016cityscapes}. 
We compared our proposed model with the state-of-the-art cross-domain detection methods. 
In this section, we present our experimental results and discussions. 

%%%%%%%%%%%%%%%%%%%%%%%%%
\subsection{Experimental Setup}
We followed the same experimental setup as in~\cite{chen2018domain}. 
The VGG16~\cite{simonyan2014very} model is used as the backbone of the Faster-RCNN detection model and pre-trained on ImageNet. 
We set the momentum as 0.9, the weight decay as 0.0005, and 
the total training epoch number as 20. 
The domain discriminator $D^l_{style}$ has three fully connected layers,
while the discriminator $D^l_{att}$ has one convolutional layer and two fully connected layers. 
For the hyperparameters involved in the proposed method, 
we set $\lambda=1$, set $\gamma$ to 5 on all blocks, and set $\varepsilon$ to 5 on block 5, and 4 on block 4. 
For all the experiments, we used the mean average precision (mAP) with a threshold of 0.5 to evaluate the results.

\begin{table*}[t]
\begin{center}
\caption{Detection results on the validation set of the Foggy Cityscapes. 
	SD and SA denote the two major components of the proposed SSA-DA:
	SD denotes style domain adaptive component, and SA denotes spatial attention enhanced feature alignment. }
\label{tab:1}
\renewcommand\arraystretch{1.1}
\scalebox{1.0}{
\begin{tabular}{c|cc|cccccccc|c}
\hline
Method                & SD                     & SA  & person & rider & car  & truck & bus  & train & mcycle & bicycle & mAP  \\ \hline
Source-only           &                       &    & 25.1   & 32.7  & 31.0 & 12.5  & 23.9 & 9.1   & 23.7      & 29.1    & 23.4 \\ \hline
BDC-Faster~\cite{saito2019strong}             &                       &    & 26.4   & 37.2  & 42.4 & 21.2  & 29.2 & 12.3   & 22.6      & 28.9    & 27.5 \\ \hline
DA-Faster~\cite{chen2018domain}               &                       &    & 25.0   & 31.0  & 40.5 & 22.1  & 35.3 & 20.2   & 20.0      & 27.1    & 27.6 \\ \hline
SC-DA(Type3)~\cite{zhu2019adapting}           &                       &    & 33.5   & 38.0  &  {\bf48.5} & 26.5  & 39.0 & 23.3  & 28.0      & 33.6    & 33.8 \\ \hline
MAF~\cite{HeMulti}                            &                       &    &  28.2   & 39.5  & 43.9  & 23.8  & 39.9 & 33.3  & 29.2      & 33.9    & 34.0 \\ \hline
SW-DA~\cite{saito2019strong}                  &                       &    &  29.9   & 42.3  & 43.5  & 24.5  & 36.2 & 32.6  & 30.0      & 35.3    & 34.3 \\ \hline
DD-MRL~\cite{kim2019diversify}                &                       &    & 30.8   &  40.5  & 44.3 & 27.2  & 38.4 & 34.5  & 28.4      & 32.2    & 34.6 \\ \hline
Dense-DA~\cite{xie2019multi}                  &                       &    & 33.2   & 44.2  & 44.8 & 28.2  & 41.8 & 28.7  & 30.5      & 36.5    & 36.0 \\ \hline
\multirow{3}{*}{SSA-DA} 
& \checkmark &    
& 33.3   & 46.2  & 44.0     & 31.1  &  47.7    & 36.4    & 36.1    &36.4    & 38.9  
\\ \cline{2-12} 
&  & \checkmark   
&32.7    &47.5   & 44.9     & {\bf36.2}   & 43.7     & 23.4    & 38.0    & 36.5   & 37.8 
\\ \cline{2-12}  
& \checkmark& \checkmark   
&{\bf33.9}    &{\bf48.3}   & 47.7     & 35.7   & {\bf52.0}     & {\bf44.7}    & {\bf39.6}    & {\bf37.9}    & {\bf42.5}     \\ \hline

\end{tabular}}
\end{center}
%\vskip -.1in	
\end{table*}

\subsection{Normal to Foggy Weather Adaptation}\label{section:4.2}

For object detection in real road scenarios, weather condition is a common factor that affects the detection performance. 
The change of weather conditions can lead to large visual variations in images and videos, 
which presents an obvious domain shift situation for the deployment of object detectors. 
To test the proposed adaptive detection model in this scenario, 
we used the cross-domain detection task from
Cityscapes~\cite{cordts2016cityscapes} 
to Foggy Cityscapes~\cite{sakaridis2018semantic}. 
These two datasets have eight object categories:
person, rider, car, truck, bus, train, motorcycle and bicycle. 
Foggy Cityscapes is a fog dataset synthesized from Cityscapes,
which can simulate the fog weather condition in real scenes. 
The Cityscapes are used as the source domain, and the training set of Foggy Cityscapes is used as the target domain. 
We set the hyperparameter  $\mu$ = 0.5 in the experiment, and report detection results for all categories on the Foggy Cityscapes validation set.

We compared the proposed SSA-DA method with seven state-of-the-art cross-domain detection methods and one baseline source-only training method.
The comparison results are reported in Table~\ref{tab:1}. 
As the baseline Source-only is only trained in the source domain without handling the domain shift problem,
we can see that all the other domain adaptive detection methods outperform Source-only. 
By having both style and spatial attention enhanced feature alignments,
the proposed SSA-DA greatly improves the cross-domain detection performance, 
far exceeding all other comparison methods. 
It outperforms the Source-only baseline by 19.1\% in terms of the average mAP. 
As the domain shift is caused by the fog  in this task,
there is a significant stylistic difference across domains. 
We can see that by using the style domain adaptive component (SD) alone in SSA-DA, 
it has already outperformed the best comparison method by 2.9\% in terms of average mAP. 
Meanwhile 
by using only the spatial attention enhanced feature alignment component (SA),
SSA-DA works very well in the `truck' category, where all the other comparison methods have poor performance. 
This suggests that the `truck' object under the condition of fog is very difficult to capture, 
while the spatial attention mechanism can help mitigate the problem.
However, excessive attention alone may have a negative impact on the category of `train', 
while the style feature alignment component can help to mitigate this drawback. 
Overall, by integrating both the SD and SA components, the proposed SSA-DA approach demonstrates great performance.  

%%%%%%%%%%%%%%%
\begin{table}[t]
\begin{center}
\caption{Detection results on adaptation from SIM-10k to Cityscapes. 
	SD and SA denote the two major components of the proposed SSA-DA:
	SD denotes style domain adaptive component, and SA denotes spatial attention enhanced feature alignment. }
\label{tab:2}
\renewcommand\tabcolsep{8pt}
\renewcommand\arraystretch{1.1}
\begin{tabular}{c|cc|c}
\hline
Method                & SD                     & SA  & AP of Car  \\ \hline
Source-only           &                        &     & 34.3    \\ \hline
BDC-Faster~\cite{saito2019strong}       &       &     & 31.8        \\ \hline
DA-Faster~\cite{chen2018domain}         &       &     & 39.0        \\ \hline
MAF~\cite{HeMulti}                      &       &     & 41.1        \\ \hline
SW-DA~\cite{saito2019strong}            &       &     & 40.1         \\ \hline
SW-DA($\gamma$=3)~\cite{saito2019strong}&       &     & 42.3         \\ \hline
SC-DA(Type3)~\cite{zhu2019adapting}     &       &     & 43.0       \\ \hline
Dense-DA(n=6)~\cite{xie2019multi}       &       &     & 42.8         \\ \hline
\multirow{3}{*}{SSA-DA} 
                            & \checkmark &               &42.0   \\ \cline{2-4} 
                            &  & \checkmark            & 42.4   \\ \cline{2-4}  
                            & \checkmark  & \checkmark    & 43.8 \\\hline

\end{tabular}
\end{center}
%\vskip -.1in	
\end{table}

%%%%%%%%%%%%%%%
\subsection{Virtual to Real Scene Adaptation}
As it is difficult to collect annotated data in many application tasks, 
it is a good option to use computer-generated labeled virtual image data for training models. 
However due to the visual difference between the virtual data and real data, 
the performance of the detection model trained on the virtual data can severely degrade when applying to the real data,
hence cross-domain detection techniques are important.
In this experiment, we tested the proposed SSA-DA method on the 
domain adaptive detection task from virtual scenes to real scenes. 
In particular, we adopted the virtual scene dataset SIM-10K~\cite{JohnsonDriving} as the source domain, 
and took the real scene dataset Cityscapes~\cite{cordts2016cityscapes} as the target domain, 
while using the car category detection as the domain adaptive detection task. 
All training images from both domains were used during training, 
test evaluation was conducted on the validation set of Cityscapes.
Following~\cite{chen2018domain},  
we set the trade-off hyperparameter $\mu$ = 0.1. 

Same as above, we compared the proposed SSA-DA with both the Source-only baseline and a number of state-of-the-art methods
which were tested on this cross-domain detection task.
The comparison results are reported in Table~\ref{tab:2}. 
We can see that our proposed SSA-DA improves the performance of the Source-only baseline model by 9.5$\%$, 
and exceeds the best results of all the other cross-domain detection models. 
It can also be observed that 
even with only one of the two components, SD and SA, SSA-DA 
can still reach a good performance level and outperform some of the latest methods.
In addition, we can also observe 
that the SC-DA(Type3) method produces the best result among the other comparison method
and it outperforms the latest Dense-DA method. 
Meanwhile, 
SC-DA(Type3) also demonstrates an obvious advantage over other comparison methods in the category `car' in 
the experiment of Section~\ref{section:4.2}. 
This validates that SC-DA(Type3) 
is more suitable for small vehicle detection.  
Nevertheless, our proposed SSA-DA outperforms SC-DA(Type3).
These results suggest that the proposed SSA-DA is very effective for cross-domain detection.

%%%%%%%%%%%%%%%
\begin{table}[t]
\begin{center}
\caption{Detection results of cross camera adaptation from KITTI and Cityscapes. 
	SD and SA denote the two major components of SSA-DA:
	SD denotes style domain adaptive component, and SA denotes spatial attention enhanced feature alignment. }
\label{tab:3}
\renewcommand\tabcolsep{8pt}
\renewcommand\arraystretch{1.1}
\begin{tabular}{c|cc|c}
\hline
Method                & SD                     & SA  & AP  of Car \\ \hline
Source-only           &                         &     & 30.2    \\ \hline
DA-Faster~\cite{chen2018domain}         &       &     & 38.5       \\ \hline
SC-DA(Type3)~\cite{zhu2019adapting}     &       &     & 42.5        \\ \hline
MAF~\cite{HeMulti}                      &       &     & 41.0       \\ \hline
\multirow{3}{*}{SSA-DA} 
                            & \checkmark &               &42.6  \\ \cline{2-4} 
                            &  & \checkmark            & 42.2   \\ \cline{2-4}  
                            & \checkmark  & \checkmark    &43.3  \\\hline

\end{tabular}
\end{center}
%\vskip -.1in	
\end{table}

%%%%%%%%%%%%%%%
\subsection{Cross-Camera Adaptation}

Due to the variations in 
camera equipments and collection scenes, real road condition data acquired under similar weather conditions 
can also have a domain shift problem. 
In this experiment, we used two real datasets, KITTI~\cite{geiger2012we} and Cityscapes~\cite{cordts2016cityscapes}, 
to study cross-domain object detection under cross-camera variations.
Following~\cite{chen2018domain}, we used the KITTI dataset as the source domain, used the Cityscapes training set as the target domain, 
and evaluated the performance of adaptive detection models on the validation set of Cityscapes with the category `car'.
The experimental results are reported in Table~\ref{tab:3}. 
We can see that the proposed SSA-DA method produced the best result, which is 13.1$\%$ higher than the baseline, 
and outperforms even the more complex SC-DA(Type3) models that are suitable for automotive inspection.

%%%%%%%%%%%%%%%%%%%%%%%%%%%%%%%%%%%%%%%%%%%%%%%%%%%

\begin{figure}[!h]
\centering
\subfigure[Cityscapes]{
\includegraphics[width=0.235\linewidth]{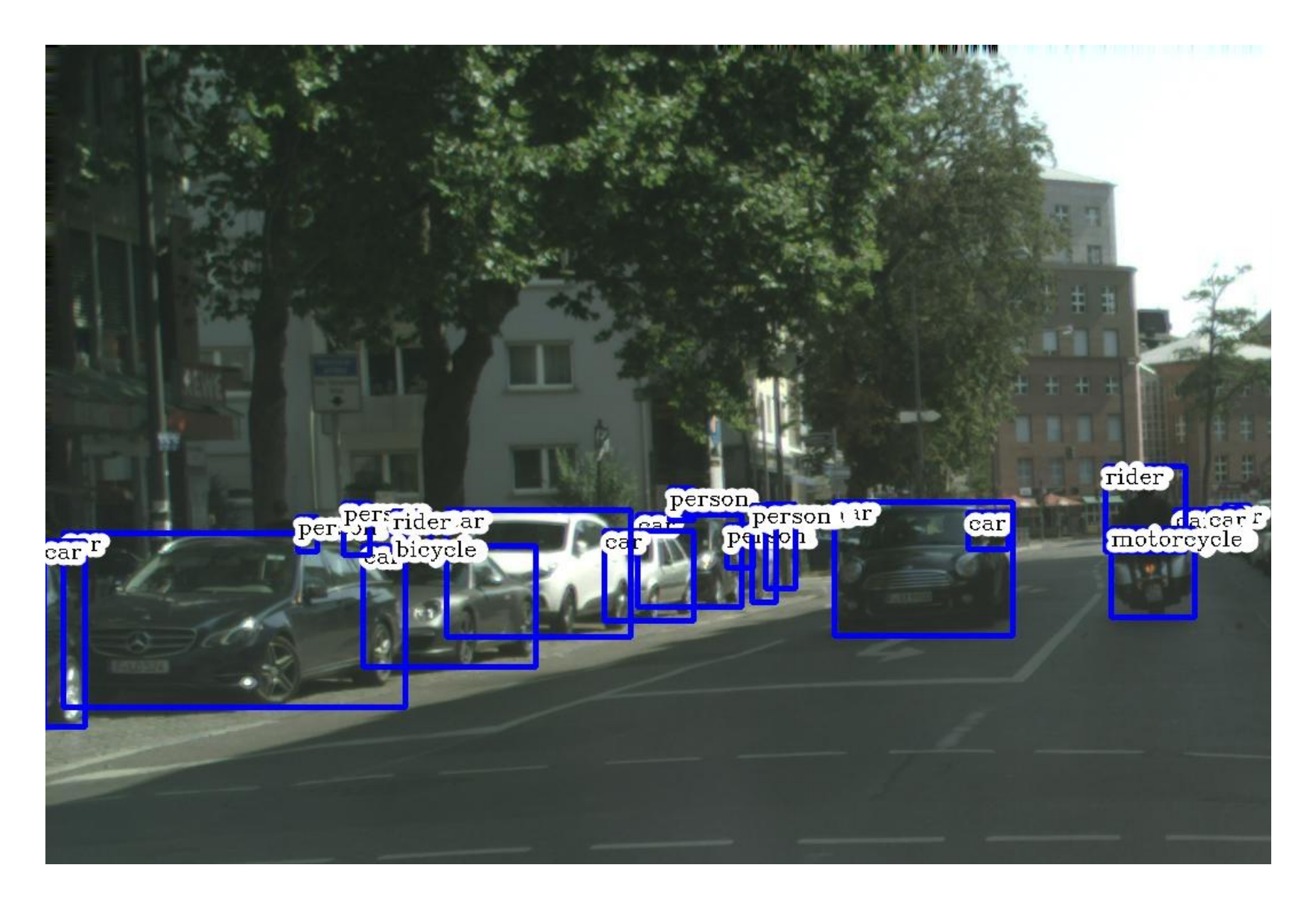}}
\subfigure[Source-only]{
\includegraphics[width=0.235\linewidth]{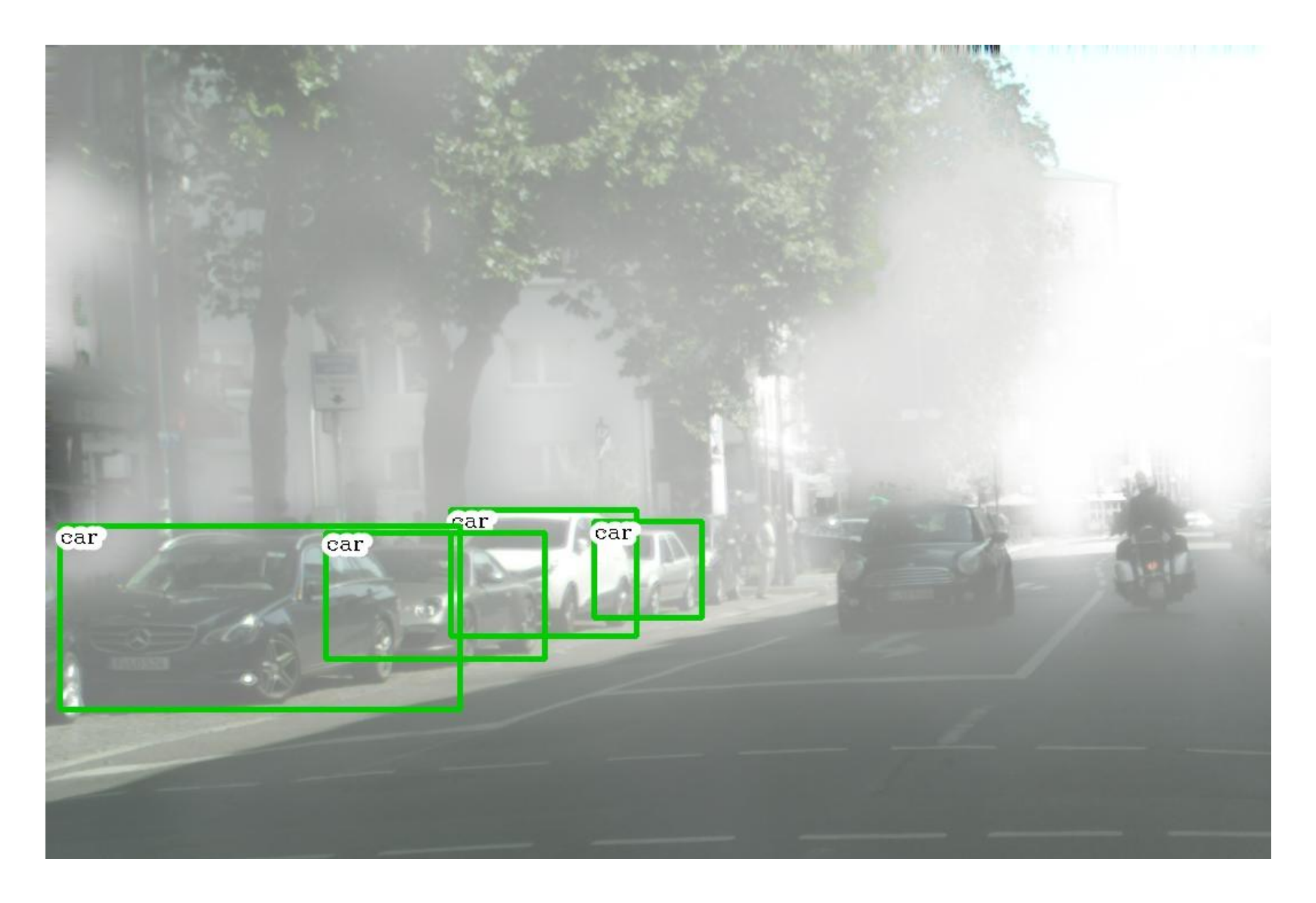}}
\subfigure[DA-Faster]{
\includegraphics[width=0.235\linewidth]{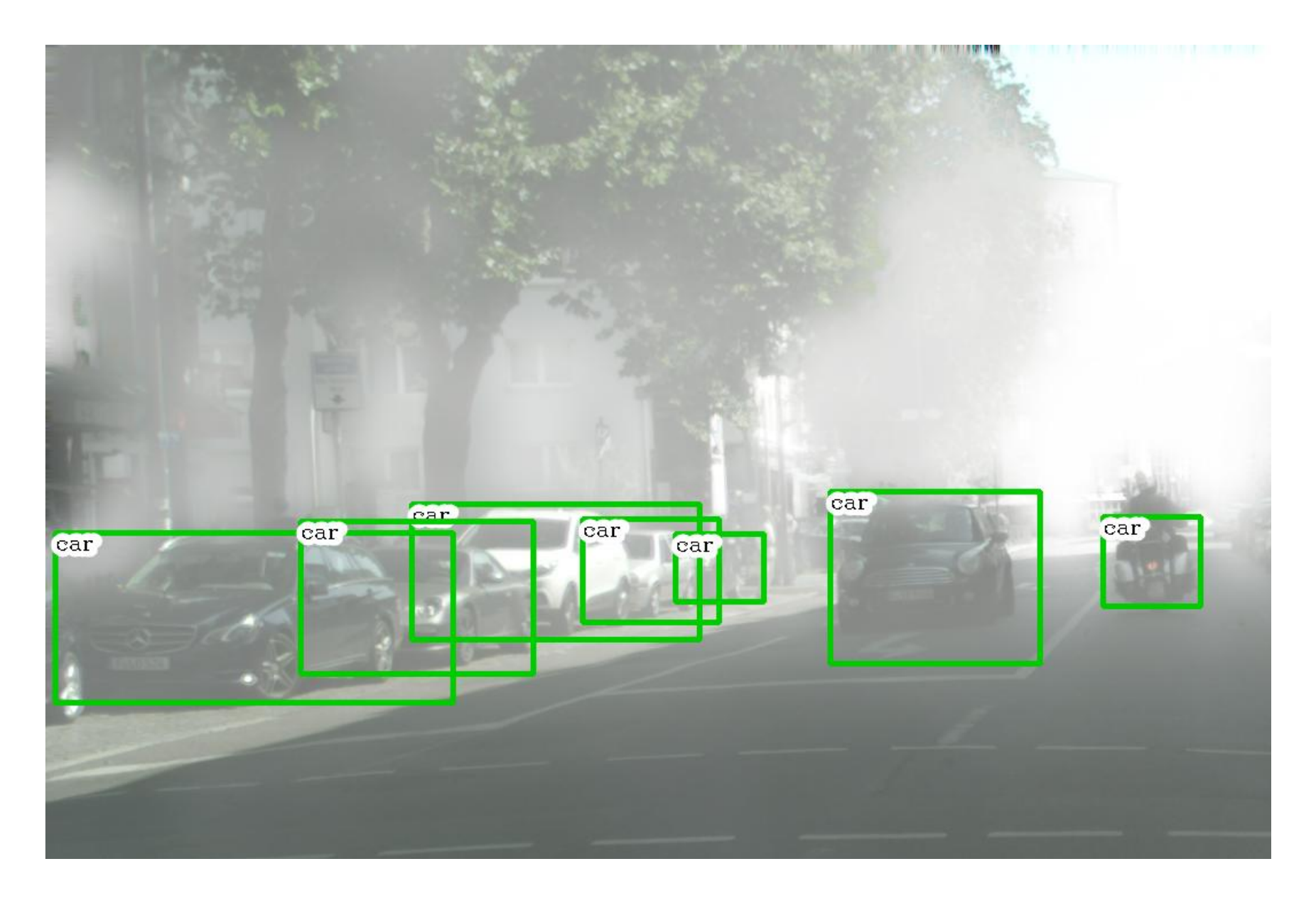}}
\subfigure[SSA-DA]{
\includegraphics[width=0.235\linewidth]{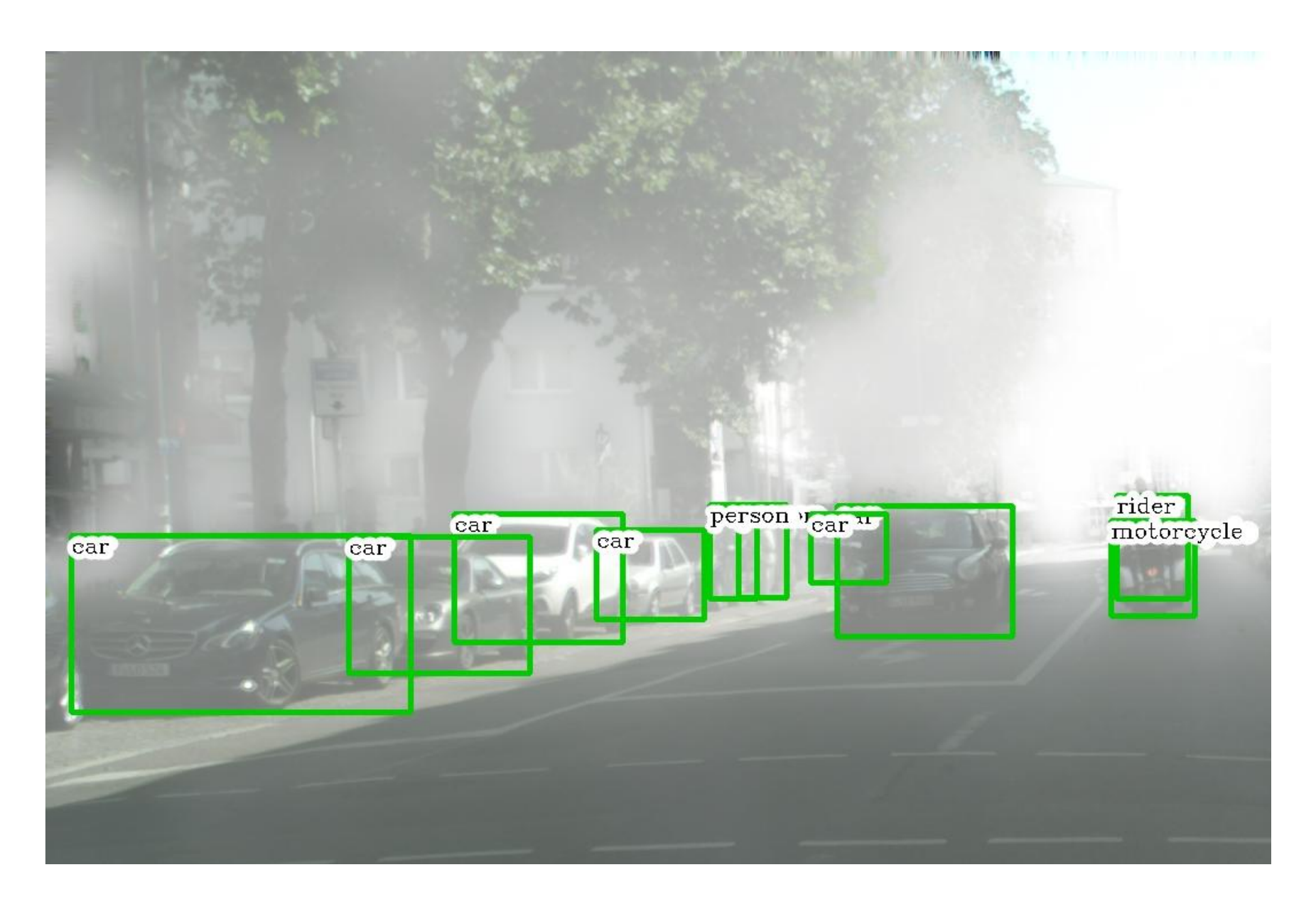}}
   \caption{Qualitative results of adaptation from Cityscapes to Foggy Cityscapes. 
	(a) Annotated image in Cityscapes. (b) Detection results of Source-only in the target domain. 
	(c) Detection results of DA-Faster. (d) Detection results of SSA-DA. 
	The blue boxes show ground-truth and the green boxes show the detection results.}
\label{fig:five}
%\vskip -.1in	
\end{figure}

\subsection{Qualitative Result Visualization}
In addition to the quantitative results reported above,
we present an example of the qualitative adaptive detection results in 
Fig.~\ref{fig:five}.  
We can see that the Source-only baseline can only detect objects within close range 
and missed most objects far away. 
DA-Faster was able to detect cars that were a little further away, 
but it mistakenly classified the motorcycle, and missed the rider, person, as well as many other objects. 
The proposed SSA-DA model correctly detected motorcycle and rider, 
and detected more person and car objects in the dense fog.

%%%%%%%%%%%%%%%%%%%%%%%%%%%%%%%%%%%%%%%%%%%%%%%%%%%%%%%%%%%

\subsection{Analysis of the SSA-DA Model}
The proposed SSA-DA model has two major components, 
the style domain adaptive module (SD) and the spatial attention enhanced feature alignment module (SA).
In this section, we will analyze the impact of these two components at multi-levels
on the performance of the SSA-DA model using the adaptive detection task from Cityscapes to Foggy Cityscapes, 
as well as investigate the impact of the hyperparameters on the components and the full model.
\\

\noindent{\bf Impact of the Component Modules.}
We use SD-DA to denote the variant of SSA-DA 
that drops the SA modules and keeps only the SD modules. 
Similarly, we use SA-DA to denote the variant of SSA-DA
that drops the SD modules and keeps only the SA modules. 
Both modules are applied on multiple blocks (block 3, 4 and 5) of the backbone detection network,
we hence further conducted ablation study to investigate their impacts on lower levels of convolution layers
and upper levels of convolution layers. 
As the style features obtained at the lower levels of the network reflect more detailed pixel level information
and the style features at the higher level reflect smooth structural information, 
we performed ablation study on SD-DA by adding the SD module from the lower level blocks to the higher level blocks. 
Semantic content information however is more prevalent at higher levels of the extraction network. 
Hence we performed ablation study on SA-DA by adding the SA module from the higher level blocks to the lower level blocks.
The experimental results are reported in Table~\ref{tab:4}. 
We can see that the overall performance of the SD-DA gradually increase by adding the SD module into higher level blocks. 
This proves that 
simultaneous style feature alignments at multiple blocks from the low level to the high level of the network
can benefit the domain adaptation performance. 
Meanwhile, for SA-DA, adding the SA module to the low level block 3 actually degrades its overall performance.
It has a significant negative impact on the `train' category.
This suggests that with spatial attention, it is more suitable to conduct semantic content feature alignment at higher levels of the network.
\\

\begin{table}[t]
\begin{center}
\caption{Detection results of SD-DA and SA-DA with deployment on different blocks.}
\label{tab:4}
\renewcommand\arraystretch{1.1}
\begin{tabular}{c|ccc|cccccccc|c}
\hline
Method                & block5    &block4       &block3  & person & rider & car  & truck & bus  & train & mcycle & bicycle & mAP  \\ \hline

\multirow{3}{*}{SD-DA} 

&  &    &\checkmark    
&32.3   &44.4      &43.7   &28.8     &42.6      &22.7     &33.3  &38.2  &35.8 \\ 
\cline{2-13} 
&  &\checkmark   &\checkmark    
& 32.5  &45.7      &43.8   & 26.8     & 49.6     &30.0     & 33.5 &37.5  &37.4 \\ \cline{2-13}  
& \checkmark&\checkmark&\checkmark   
	& 33.3   & 46.2  & 44.0     & 31.1  &  47.7    & 36.4    & 36.1    &36.4    & {\bf 38.9}  
\\ \hline
\multirow{3}{*}{SA-DA} 
& \checkmark &    &    
&30.6   &40.0     & 40.5  & 26.4    &  37.4    & 27.3    & 30.0     & 33.4   & 33.2  
\\ \cline{2-13} 
&\checkmark &\checkmark  & 
	&32.7    &47.5   & 44.9     & 36.2   & 43.7     & 23.4    & 38.0    & 36.5   & {\bf 37.8}  
\\ \cline{2-13}  
& \checkmark&\checkmark&\checkmark   
&33.9    &47.2    & 44.9     &35.5     &41.2   & 11.9  &40.0   &35.6    & 36.3
\\ \hline
\end{tabular}
\end{center}
%\vskip -.2in	
\end{table}
%%%%%%%%%%%%%%%%%%%

\noindent{\bf Parameter Sensitivity on $\gamma$ and $\varepsilon$}. 
These two parameters are modulation factors for the focal loss 
used in the adversarial alignment of style features (SD module)
and spatial content features respectively (SA module).
As they are separately involved in the two modules, 
we conducted sensitivity experiments for $\gamma$ and $\varepsilon$ using
SD-DA and SA-DA respectively.

\begin{figure}[t]
\centering
\subfigure[]{
\includegraphics[width=0.29\linewidth]{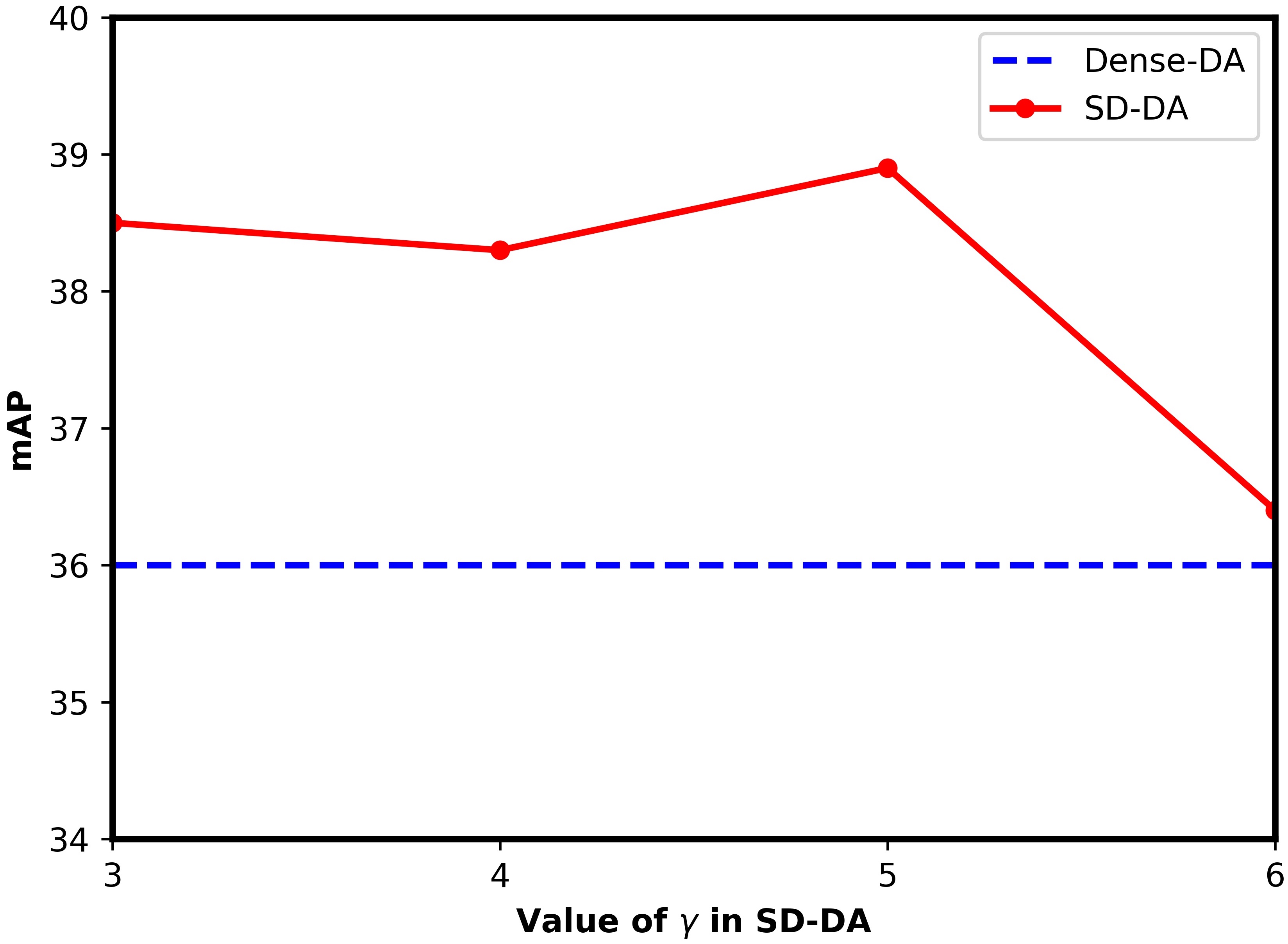}}
\quad
\subfigure[]{
\includegraphics[width=0.29\linewidth]{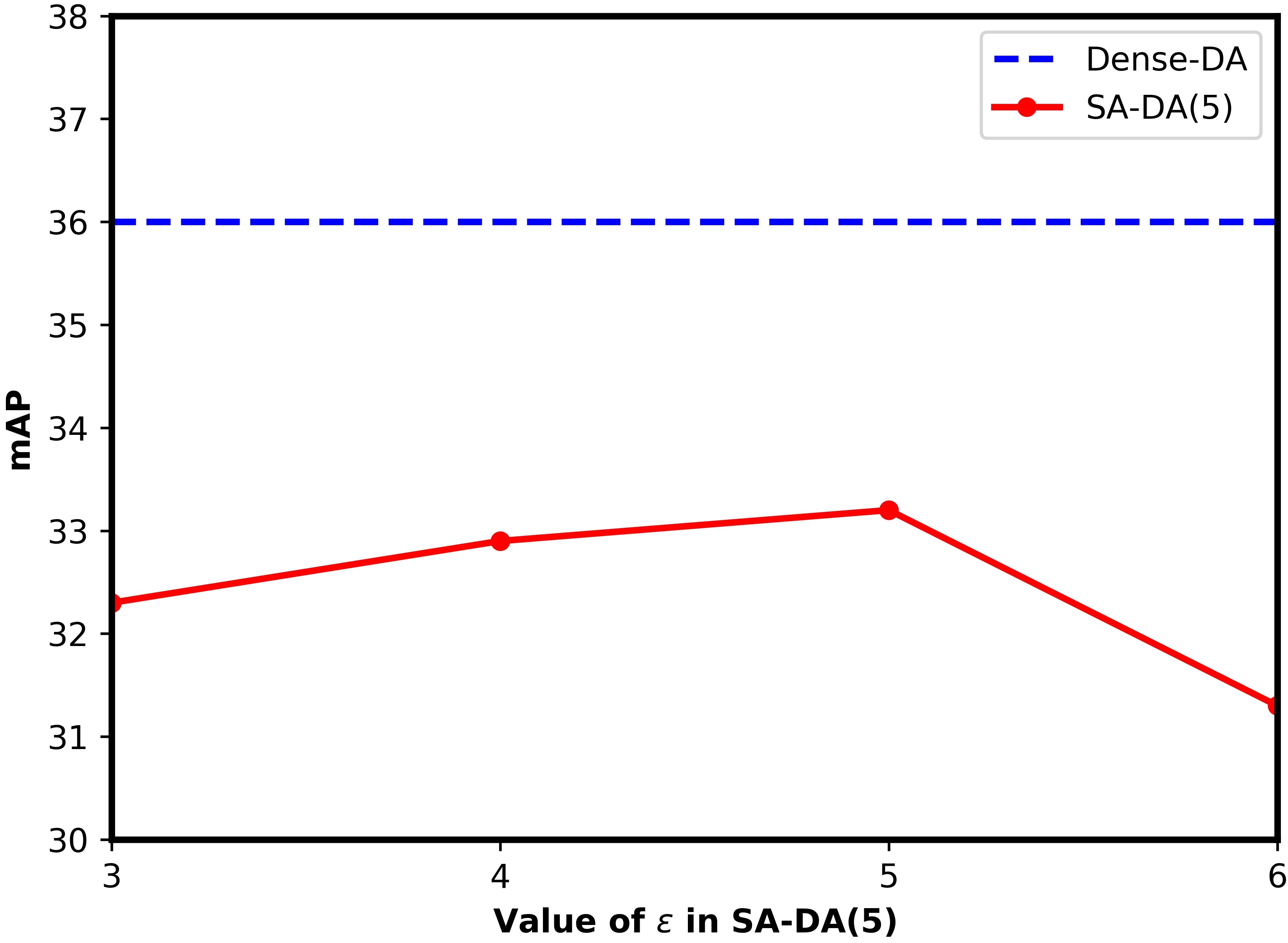}}
\quad
\subfigure[]{
\includegraphics[width=0.29\linewidth]{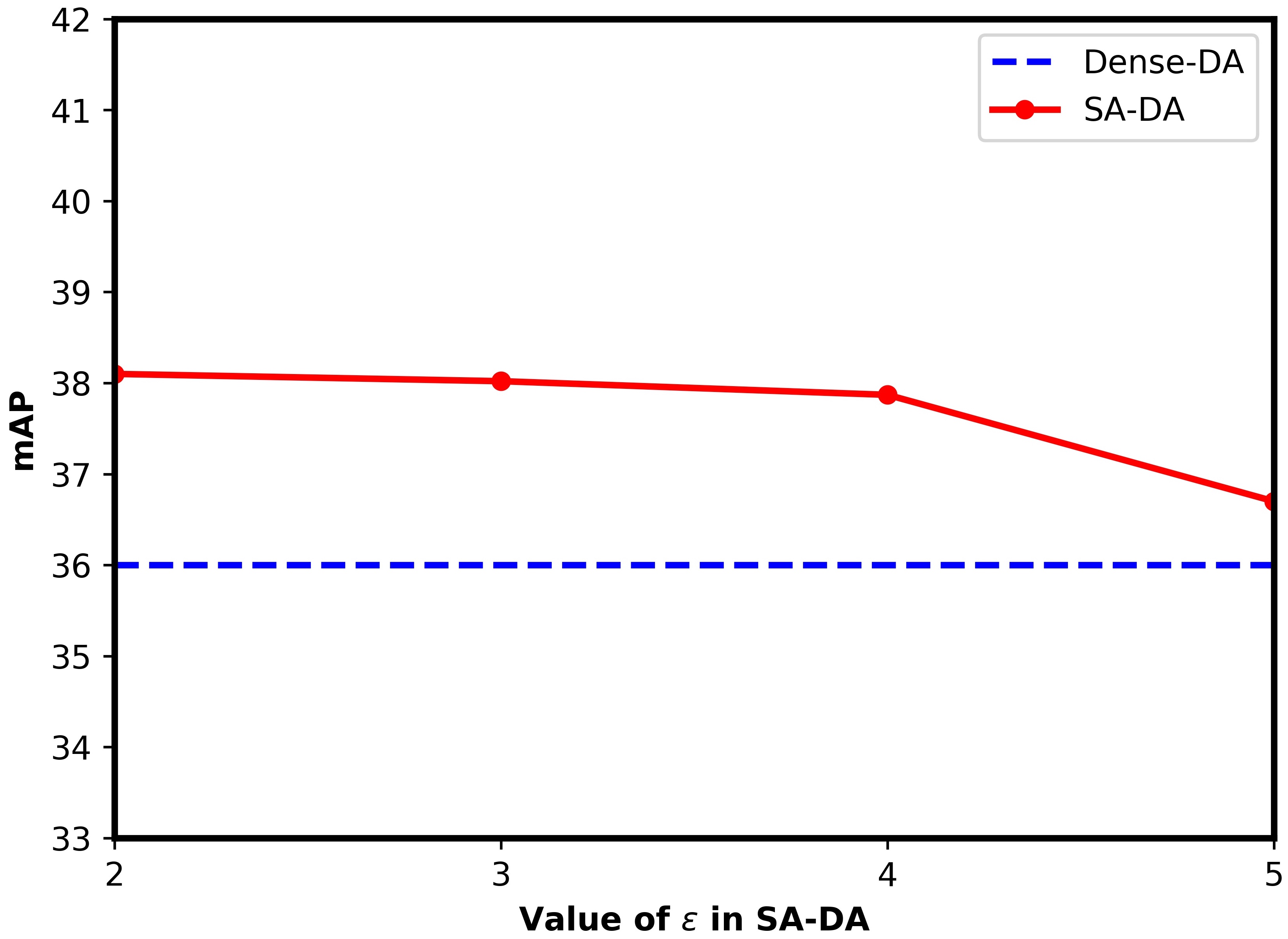}}
%\vskip -.1in	
   \caption{Parameter sensitivity on $\gamma$ and $\varepsilon$ with adaptation from Cityscapes to Foggy Cityscapes. 
{(a)}: Sensitivity results over $\gamma$ with SD-DA. 
{(b)}: Sensitivity results over $\varepsilon$ with SA-DA(5).
{(c)}: Sensitivity results over $\varepsilon$ with SA-DA.}
\label{fig:three}
\end{figure}

As the style feature alignment has been shown to be useful at both low and high level blocks,
we set $\gamma$ to the same value for the SD modules added to block 3, 4 and 5.
We tested the SD-DA variant by varying the $\gamma$ within the range of values $[3, 4, 5, 6]$,
and the results are reported in Fig.~\ref{fig:three}(a).
We can see SD-DA outperforms Dense-DA for different $\gamma$ values, especially when $3\leq \gamma\leq 5$, 
while the best result is achieved with $\gamma=5$.

From previous experiments, we can see that the SA module is more suitable for higher level blocks.
Hence here we separately investigated the best $\varepsilon$ value for the SA module used in block 4 and block 5.
First we use a variant, SA-DA(5), which only adds SA module to block 5, to conduct sensitivity experiments
by varying $\varepsilon$ within the range of values $[3, 4, 5, 6]$.
The results are reported in Fig.~\ref{fig:three}(b).
We can see adding SA module only to block 5 leads to degraded results comparing to the full SA-DA.
The performance varies with different $\varepsilon$ values while the best result is gained with $\varepsilon=5$.
Then we fixed $\varepsilon=5$ for the SA module in block 5, 
and tested the full SA-DA method 
by varying $\varepsilon$ in block 4 within the range of values $[2, 3, 4, 5]$.
The results are reported in Fig.~\ref{fig:three}(c). 
We can see that when $\varepsilon$ = $\{2,3,4\}$ on block4, 
the performance remains at a high level, but will drop when $\varepsilon$ continues to increase.
Overall, these results suggest $\gamma$ and $\varepsilon$ should not be set to big values
as they may overfit the hard examples. A value no larger than 5 would be more suitable.
\\

%%%%%%%%%%%%%%%%%%%
\begin{figure}[t]
\centering
\subfigure[]{
\includegraphics[width=0.33\linewidth]{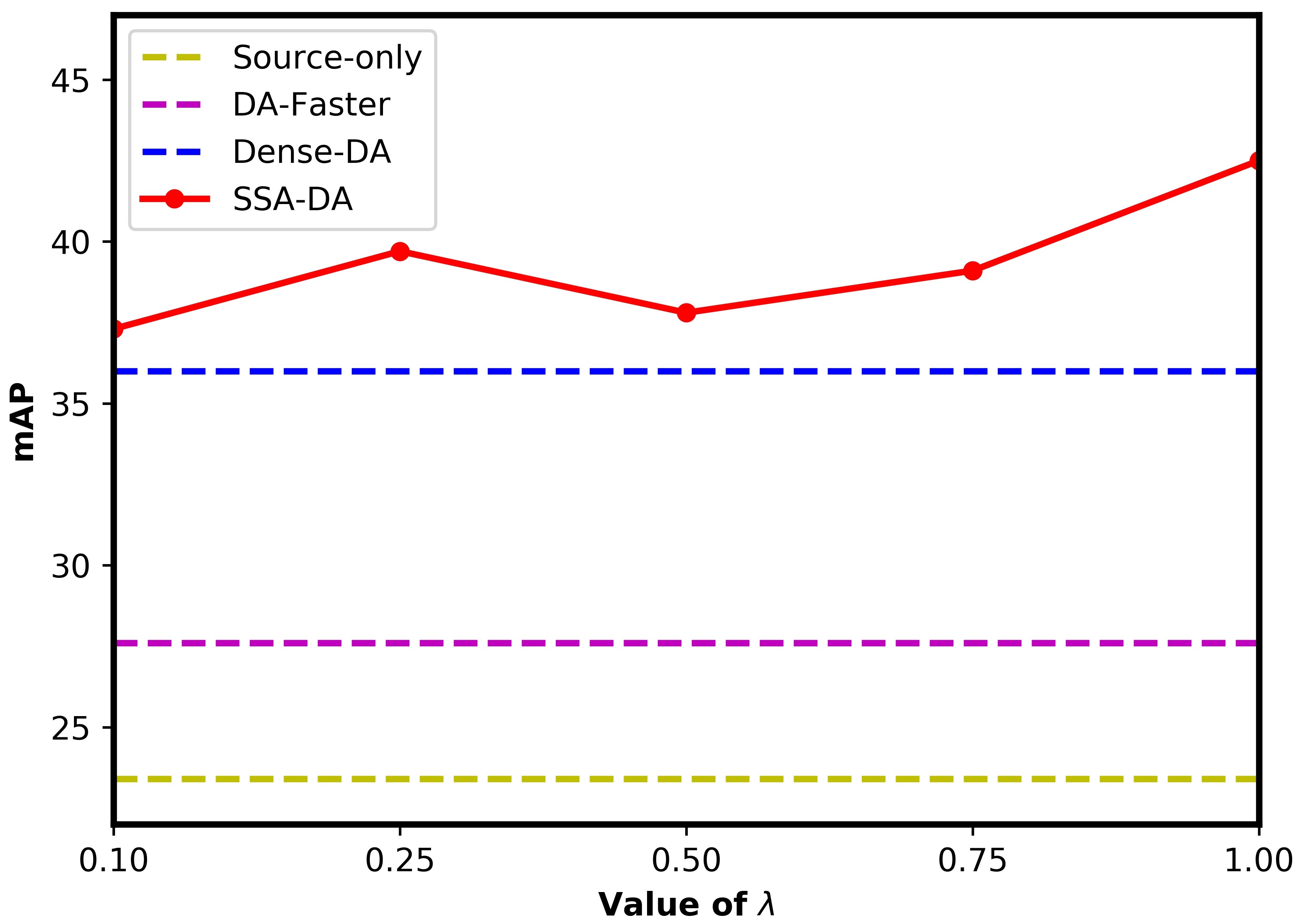}}
\quad
\subfigure[]{
\includegraphics[width=0.33\linewidth]{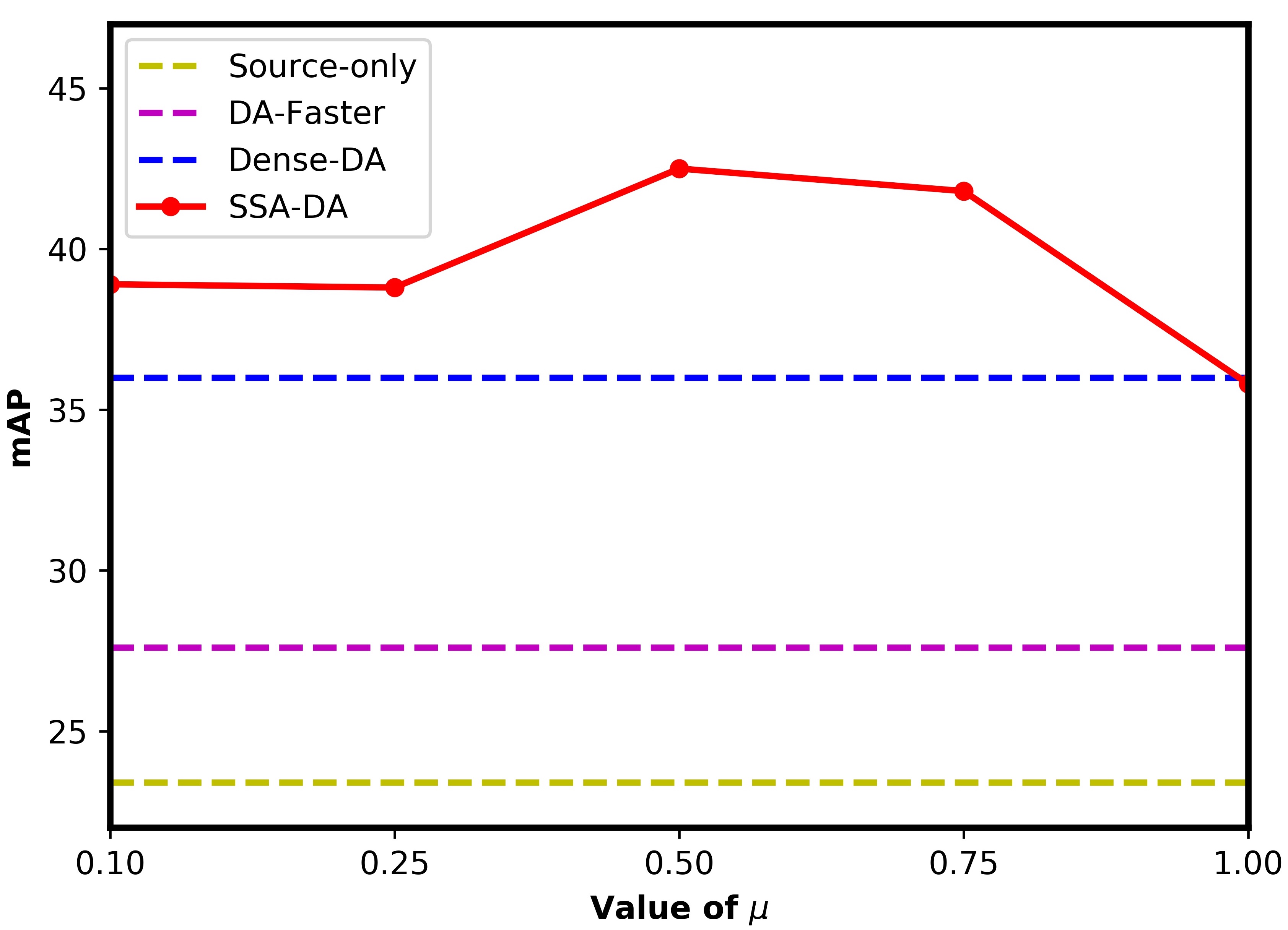}}
   \caption{Parameter sensitivity analysis on $\lambda$ and $\mu$ with the adaptation from Cityscapes to Foggy Cityscapes. 
	{(a)}: Sensitivity results over $\lambda$. {(b)}: Sensitivity results over $\mu$.}
\label{fig:four}
%\vskip -.1in	
\end{figure}
%%%%%%%%%%%%%%%%%%%
\noindent{\bf Parameter Sensitivity on $\lambda$ and $\mu$}. 
In addition, we conducted experiments on the complete model SSA-DA by adjusting the trade-off parameters $\lambda$ and $\mu$ in Eq.(\ref{eq11}). 
These two parameters control the weight of style domain adaption (SD) and spatial attention enhanced feature alignment (SA) 
in the detection model. 
To vary the $\lambda$ value, we fixed $\mu=0.5$; to vary the $\mu$ value, we fixed $\lambda=1$.
The sensitivity results are reported in Fig.~\ref{fig:four}.
We can see that although the performance varies with different $\lambda$ and $\mu$ values,
the performance in general is superior to Dense-DA and DA-Faster for most of range of values,
while the best results are obtained when $\lambda$=1 and $\mu$=0.5.

%%%%%%%%%%%%%%%%%%%%%%%%%%%%%%%%%%%%%%%%
\section{Conclusion}

In this paper, we proposed a novel style and spatial attention enhanced bi-dimensional feature alignment method for domain adaptive detection (SSA-DA).
The proposed method deploys two important modules, the style domain adaptive module and the spatial attention enhanced domain alignment module, at multi-levels to align features in both the depth and spatial dimensions across domains. 
With both the style and spatial attention enhanced content feature alignments, the detector trained in the source domain
can be more adaptive to the target domain.
We conducted experiments on benchmark datasets in three different cross-domain variation scenarios.
The experimental results demonstrated the proposed model achieved the state-of-the-art adaptive detection performance. 

%%%%%%%%%%%%%%%%%%%%%%%%%%%%%%%%%%%%%%%%

%\clearpage
% ---- Bibliography ----
%
% BibTeX users should specify bibliography style 'splncs04'.
% References will then be sorted and formatted in the correct style.
%
\bibliographystyle{splncs04}
\bibliography{egbib}
\end{document}